\documentclass{article}

\usepackage{arxiv}
\usepackage{subcaption}
\usepackage{graphicx}
\usepackage{cite}
\usepackage{amsmath,amssymb,amsfonts}
\usepackage{algpseudocode}
\usepackage{algorithm}

\usepackage{graphicx}
\usepackage{textcomp}
\usepackage{xcolor}
\usepackage{placeins}
\usepackage[utf8]{inputenc} 
\usepackage[T1]{fontenc}    
\usepackage{hyperref}       
\usepackage{url}            
\usepackage{booktabs}       
\usepackage{amsfonts}       
\usepackage{nicefrac}       
\usepackage{microtype}      
\usepackage{cleveref}       
\usepackage{lipsum}         
\usepackage{graphicx}
\usepackage{doi}

\title{A template for the \emph{arxiv} style}

\renewcommand{\shorttitle}{\textit{arXiv} Template}

\hypersetup{
pdftitle={A template for the arxiv style},
pdfsubject={q-bio.NC, q-bio.QM},
pdfauthor={David S.~Hippocampus, Elias D.~Striatum},
pdfkeywords={First keyword, Second keyword, More},
}

\begin{document}
\title{Dynamic Template Selection Through Change Detection for Adaptive Siamese Tracking}

\author{Madhu Kiran\thanks{Laboratoire d’imagerie, de vision et d’intelligence artificielle (LIVIA), Ecole de technologie
superieure, Montreal, Canada}\thanks{Corresponding author, madhu\_sajc@hotmail.com} \and Le Thanh Nguyen-Meidine\footnotemark[1] \and Rajat Sahay\thanks{Vellore Institute of Technology, Vellore} \and Rafael Menelau Oliveira E Cruz\footnotemark[1] \and Louis-Antoine Blais-Morin\thanks{Genetec Inc.} \and Eric Granger\footnotemark[1]}

\maketitle

\begin{abstract}
Deep Siamese trackers have recently gained much attention in recent years since they can track visual objects at high speeds. Additionally, adaptive tracking methods, where target samples collected by the tracker are employed for online learning, have achieved state-of-the-art accuracy. However, single object tracking (SOT) remains a challenging task in real-world application due to changes and deformations in a target object's appearance. Learning on all the collected samples may lead to catastrophic forgetting, and thereby corrupt the tracking model.
In this paper, SOT is formulated as an online incremental learning problem. A new method is proposed for dynamic sample selection and memory replay, preventing template corruption. In particular, we propose a change detection mechanism to detect gradual changes in object appearance, and select the corresponding samples for online adaption. In addition, an entropy-based sample selection strategy is introduced to maintain a diversified auxiliary buffer for memory replay. Our proposed method can be integrated into any object tracking algorithm that leverages online learning for model adaptation.
Extensive experiments conducted on the OTB-100, LaSOT, UAV123, and TrackingNet datasets highlight the cost-effectiveness of our method, along with the contribution of its key components. Results indicate that integrating our proposed method into state-of-art adaptive Siamese trackers can increase the potential benefits of a template update strategy, and significantly improve performance.
%
\end{abstract}


\begin{figure}[b!]
 \centering
\includegraphics[width=1.0\linewidth]{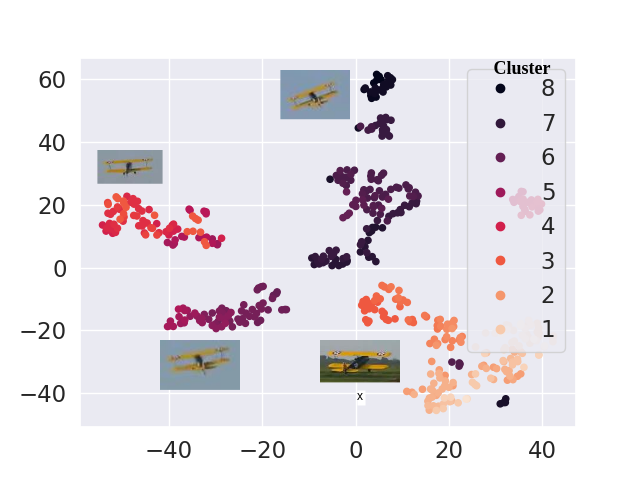}
   \caption{\textbf{Motivation of our proposed method}: A t-SNE plot of deep CNN features vectors extracted from an object bounding boxes during tracking. A vector was sampled every time a gradual change was flagged by a change detector. Similar points indicate similar appearance, while similar colors indicate vectors sampled close in time, until a more abrupt change is detected (see the legend). Although vectors sampled between more abrupt changes tend to be clustered, they may also overlap with those extracted at a different time of change.}
  \label{fig:tsne}. 
  \vspace{-0.35cm}
\end{figure}

\section{Introduction}
\label{submission}

Along with the success of deep learning (DL) models~\cite{bertinetto2016fully, SiamRPN,siamtriplet,DaSiam,SiamDW}, several object trackers have adopted online learning~\cite{nam2016mdnet,eco,atom,dimp,prdimp,kws} to improve performance. These trackers can be categorized as adaptive tracking methods (or tracking-by-detection methods), and template matching methods~\cite{bertinetto2016fully,DaSiam,SiamDW,SiamRPN}. Adaptive Siamese tracking methods learn the target model online either through stochastic gradient descent or other methods of optimization, such as steepest gradient descent~\cite{dimp,prdimp,kws,transdimp}. They learn an object classifier online to distinguish an object from the background. Template matching methods compare an initial object appearance template with a search region. They handle change in object appearance by interpolating the initial template with the recently generated object template, as in~\cite{correl, updtnet}. 

Some early DL models for trackers, such as MDNet and TCNN~\cite{nam2016mdnet,nam2016modeling}, rely on computationally expensive online model adaptation methods by gradient descent. Despite the high level of tracking accuracy, they were not applicable for real-time scenarios. Recent trackers, such as~\cite{eco,atom,dimp,prdimp,kws}, have introduced efficient optimization methods for online learning of classifiers given a memory of previously seen templates localized by the tracker. Given the ability to efficiently update these classifiers in response to changing object appearance, adaptive Siamese trackers generally provide higher accuracy than template matching. 

In this paper, we focus on the problem of online model adaptation in adaptive Siamese trackers, or tracking-by-detection methods. Some of the main challenges with object tracking, and in particular with adaptive Siamese trackers are: (1) object appearance changes over time, can be recurrent or a permanent change; (2) model adaptation with template generated by the tracker can be noisy, and during online training of the classifier, these are considered as samples with pseudo labels; and (3) rapid model adaptation is required to address real-time applications. Most adaptive Siamese trackers update the classifier after some basic sanity check. Frequent classifier training leads to integration of noise from tracker localized samples, and thereby the overall model adaptation is not efficiently utilized. 

Figure~\ref{fig:tsne} shows the t-SNE plot of bounding boxed sampled from a LaSOT~\cite{lasot} video, where each sample is represented using features extracted by the DiMP~\cite{dimp} tracker. Assume that a change detector~\cite{agrahari2021concept} is applied on the similarity matching scores to assess concept drift\footnote{We define concept drift as a change in the distribution of data over time in online learning~\cite{agrahari2021concept}. Gradual, recurrent, and abrupt changes are more relevant in this work (see definitions in Section 2.C).} on template samples localized by the tracker. Similar colors indicate vectors sampled close in time, while similar points correspond to templates of the object that have a similar appearance, between two detected gradual changes. Given the more abrupt changes in object appearance over time, multiple modes emerge in the object feature distribution. Between two consecutive abrupt changes, object template features generally appear to cluster together. However, some points of different colors overlap, indicating recurring concepts in the object appearance space. 
 
In this paper, a SOT framework is proposed to perform online-incremental learning for model adaptation. Based on Figure~\ref{fig:tsne}, we hypothesize that object appearances of consecutive samples do not change much until a more abrupt concept drift occurs. Additionally, the changes can be recurring, indicating that the object's appearance can return to one of the earlier appearances. A tracker should only adapt online when a concept drift is detected, to prevent the tracker from adapting frequently, and integrating noisy samples into the system, even when the adaptation is unnecessary. Adapting the model only in the presence of drift reduces the overall complexity by adapting the classifier fewer times. 

Most recent adaptive Siamese trackers~\cite{dimp,prdimp,kws} rely on a budgeted memory to train the classifier online with a limited number of samples, resulting in low overall complexity. However, drifting object appearances, as discussed above, may lead to a corrupted tracking model due to catastrophic forgetting of the past, as the memory limits the storage of previously seen templates. Catastrophic forgetting can efficiently be addressed in online learning by using a replay buffer, as described in ~\cite{icarl,GEM,entropyss,gradientSS}. These methods have been initially proposed for online multi-class continual learning problems. Inspired by the entropy maximization-based sampling~\cite{entropyss} we extend the algorithm to our problem, i.e., single-class incremental learning with pseudo labels. In particular, we propose a classifier score discretization-based algorithm to maintain an auxiliary memory, and alleviate the catastrophic forgetting that may occur during online tracking. 

In this paper, the overall performance of online learned SOT is improved in two respects. First, we hypothesize and set up the online learning problem in tracking as an incremental learning problem with catastrophic forgetting with change detection. Secondly, we propose a classifier score discretization-based algorithm to sustain an auxiliary memory of diverse samples from the previous time frames, and limit knowledge corruption. We integrate our proposed method into several state-of-the-art adaptive Siamese tracker, including DiMP~\cite{dimp}, PrDimp~\cite{prdimp}, and TrDiMP~\cite{transdimp}, and evaluated on the OTB video dataset~\cite{otb,lasot,uavdata,trackingnet} to demonstrate the effectiveness of our proposed method. Performance improves a margin of 2\% on these datasets, without resorting to additional parameters or change in CNN architectures and with lower overall complexity.
\begin{figure*}[h!]
 \centering
\includegraphics[width=0.99\linewidth]{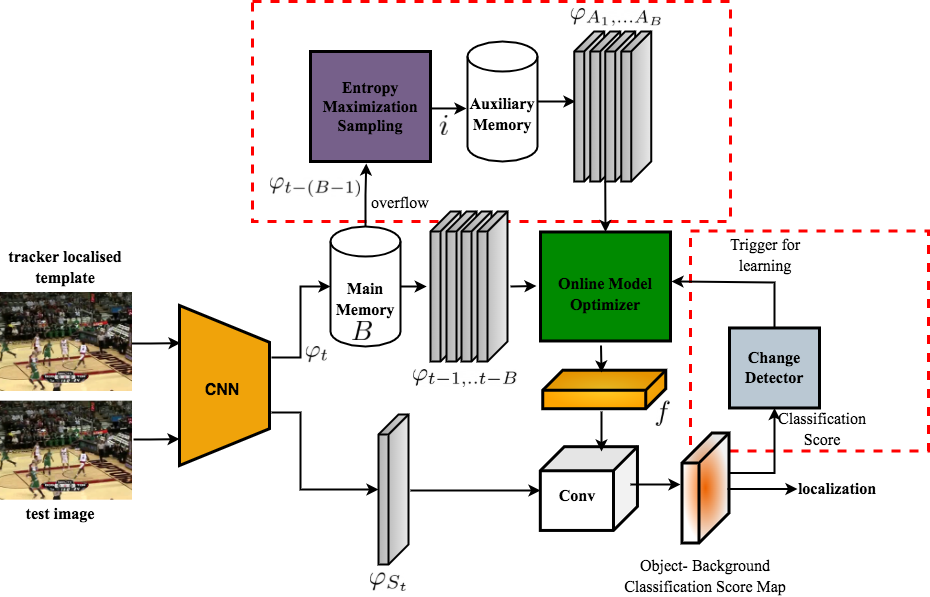}
   \caption{Overall framework of our proposed method applied to an adaptive Siamese tracker. It relies on a change detection mechanism to trigger adaptation of the model, and an additional auxiliary memory to store diverse prior template information, and  alleviate catastrophic forgetting in tracking. The regions indicated in dotted lines point to our contributions in this paper. }
  \label{fig:proposed}. 
  \vspace{-0.35cm}
\end{figure*}



\section{Related Work}

\subsection{Siamese Tracking:} 
Pioneered by SINT~\cite{sint} and SiamFC~\cite{bertinetto2016fully}, the Siamese family of trackers evolved from Siamese networks, trained offline through metric learning for similarity matching. These networks were trained on a large dataset to learn generic feature representations for object tracking. SiamRPN~\cite{SiamRPN} further improves on the pioneering work by employing region proposals to produce a target-specific anchor-based detector. Then, the following Siamese trackers mainly involved designing powerful backbones~\cite{SiamDW, SiamVGG}, or more powerful proposal networks, like in~\cite{cascaded}. ATOM~\cite{atom} and DIMP~\cite{dimp} are robust online trackers that differ from the general offline Siamese trackers. Other paradigms of Siamese trackers are distractor-aware training and domain-specific tracking~\cite{DaSiam,SA-Siam}.

\subsection{Online Learning for Tracking:} 
In \cite{reinforcement}, an LSTM is incorporated to learn long-term relationships during tracking and turns the VOT problem into a consecutive decision-making process of selecting the best model for tracking via reinforcement learning~\cite{regularity1}. In~\cite{correl} and~\cite{DaSiam}, models are updated online by a moving average based learning. These methods integrate the target region extracted from tracker output into the initial target. In~\cite{generative1}, a generative model is learned via adversarial learning to generate random masks that produce shifted versions of target templates from the original template. Then, an adversarial function is used to decide whether or not the generated template is from the same distribution and if they will be used as templates for tracking. In~\cite{duynamicmem}, an LSTM is employed to estimate the current template by storing previous templates in a  memory bank. In~\cite{dynamic}, authors propose to compute transformation matrix with reference to the initial template, with a regularised linear regression in the Fourier domain. Finally, in~\cite{yao2018joint}, authors propose to learn the updating co-efficient of a correlation filter-based tracker using SGD online. All these methods use the tracker output as the reference template while updating on top of the initial template. ~\cite{dimp,prdimp} propose a model where an adaptive discriminative model is generated online by the steepest gradient descent method. They differ from another online learned method like ~\cite{nam2016mdnet} due to their real-time performance. Similarly~\cite{ocean} introduce online model prediction but employ a fast conjugate gradient algorithm for model prediction. Foreground score maps are estimated online, and the classification branch is combined by weighted addition. All these trackers update the template every frame after some sanity check of the templates. ~\cite{metaupdater} is somewhat close to our work by the fact that they consider when a how to update the  tracker. However, it involves a computationally expensive model -- runs at a slow speed of 13fps --which is trained to decide whether the tracker is ready for an update.

\subsection{Online Incremental Learning:} 
Learning new classes or distributions of existing classes online with sequential data is called incremental learning~\cite{entropyss,he2020incremental}. In such scenarios, models suffer from poor performance on previous tasks caused by knowledge corruption of catastrophic forgetting. Regularization methods have employ constraints to learning of new tasks via a suitable regularization term~\cite{ewc,synaptic}. Structural methods freezing weight and expand the architecture along with new tasks~\cite{rusu2016progressive,dynamiclearning}. Rehearsal methods use some sort of memory to alleviate catastrophic forgetting by storing previously seen examples. Gradient episodic memory (GEM) uses examples from previously seen task and during training of a new task forces the gradient descent steps such that they don't increase the loss on them. Generative replay~\cite{generativereplay} learns a Generative adversarial network (GAN) during training of previous tasks to implicitly store samples of older tasks and mix them in appropriate proportions while training of new tasks. Gradient based Sample Selection (GSS)~\cite{gradientSS} maintains a buffer of diverse samples suing gradient information.~\cite{entropyss} use a two step method to maintain a buffer of diverse samples for a multi-class incremental learning problem using entropy. Their first step maintains a set of samples with diverse class labels. Inspired from their method, be adapt it our single class incremental learning problem.  

\subsection{Change Detection in Streaming Data:} 
 Concept drift can be categorised as gradual change, abrupt change, incremental change and recurrent change. In gradual change the duration of change in concept is relatively large compared to abrupt drift. In Recurrent drift concepts re-appear later after a drift and in incremental drift the concepts continuously change over time and they are also gradual changes. Drift detection methods can be categorised as 1) S
 equential, 2) Adaptive Windowing 3) Fixed Cumulative windowing 4) Statistical. Sequential analysis based methods predict the drift sequentially such as CUSUM~\cite{cusum} and Page-Hinckeley~\cite{phtest}. Window based methods use a fixed or adaptive window to summarize old information and newer ones to asses the difference between them ex: ADWIN~\cite{ADWIN}, HDDM~\cite{hddm}. Statistical methods use statistical parameters to to predict the drift DDM~\cite{gama2004learning}, RDDM~\cite{RDDM}. Some of the Neural network based model includes ~\cite{spiking, 3layerdrift} both designed for drift detection in text data streams. These included a neural network model for drift detection. A few drift detection algorithms were presented for unlabelled data stream drift detection. This includes ~\cite{unlabelleddrift1} which use margin density for drift detection and ~\cite{unlabeeleddrift2}, using k-means clustering and Page-Hinkeley test.

\section{Proposed Method}

Figure~\ref{fig:proposed} shows the framework of our proposed method integrated with an adaptive Siamese tracker. The tracker accepts localised object template, and then uses the feature extractor $\Theta$ to produce the features $\varphi{t}$. These features are then stored in the main memory --  a buffer to hold $B$ instances of object template features from previous frames. $B$ is the budget of the memory buffer. Once the memory budget is met, older samples are removed on a first in, first out basis. An online classifier is trained by the optimizer using all the instances from the memory buffer to produce a model $f$ which is convolved (Conv in Fig~\ref{fig:proposed}) with test image features $\varphi_{S_{t}}$ to obtain a classifications score map $S$. This score map distinguishes the object foreground from the background for localization. 

The framework discussed above is a common among some of the recent trackers such as~\cite{dimp,prdimp,kws,transdimp}. In this work, we propose using a change detector and training the model only when a concept drift is detected, instead of the conventional method of training the model every frame. Additionally, to alleviate the problem of concept drift, we propose to maintain an auxiliary memory to store older samples. This memory is sustained with a budget B, and once the memory is full, older samples are replaced with newer ones based on an entropy maximization algorithm~\cite{entropyss}. This increases the overall entropy of the samples in the memory.

\subsection{Change Detection:}
 Given a time period $[0, t]$, a set of samples denoted $S_{0, t}=\left\{d_{0}, \ldots, d_{t}\right\}$ where $d_{i}=\left(X_{i}, y_{i}\right)$ is one instance of data with features $X_{i}$ and label $y_i$, and $S_{0, t}$ follows a certain distribution which is $P_{t}(X, y)$.Concept drift can occur due to change in class conditional probability $P_{t}(X| y)$, change in class distributions $P(y)$ or appearances of new classes, or change in posterior probability $P(c|X)$. In general concept drift occurs when $P_{t}(X, y) \neq P_{t+1}(X, y)$. At the beginning, the tracker is initialized by a ground truth template representing the object. However, since the object's appearance may change over time, the tracker must be updated with object samples collected by the tracker (which is prone to noise) while tracking. The trackers in the literature updates the model periodically with a fixed period. However, we hypothesize that a tracker needs to be updated only when a concept drift is detected to avoid integrating noise into the model. Hence we propose using the same model initially trained at the start of the tracking until a concept drift is detected. The classification score map is a good indicator of change in concept as the distribution of incoming target samples changes the old model will perform poorly on these samples.  Since the classifier is a binary classifier, the classification score directly gives an estimation of probability of the predicted label $y$ for a given object localised by the tracker. Hence concept drift which can be detected by the joint probability of $P(\varphi{t},y)$, the classifier change is a good estimate of concept change. We employ an off-the-shelf available sequential change detector (see Section 2) that relies on the maximum value of the classification score map. The change detector statistically analyses the classification scores as a time series data and triggers a flag to indicate the change in classification score. This event triggers the classifier's training online to obtain a new model. This tracker will be used for tracking until the next change is observed.  

\subsection{Entropy Maximization Sampling for Auxiliary Memory:}
We propose an auxiliary memory of buffer size $B$ to hold older samples discarded by the main memory. The auxiliary memory will contain some of the older representations of the object in order to alleviate the problem of catastrophic forgetting due to the limited capacity of the main memory. Training the classifier with a large number of samples can affect the tracker's overall run time, making it inapplicable for real-time scenarios. Hence, it is important to maintain a limited set of samples with maximum variance to contain object representations from different time frames. We consider the problem of adaptive Siamese tracking as a variation of incremental learning from pseudo labels (labels generated by the tracker for the incoming data). As discussed in Section 2, most of the previous work in this field was related to multi-class incremental learning. Hence we propose a method to adapt the problem to binary classification. 
Initially, the auxiliary memory is filled by samples retrieved from the main memory by First in First Out (FIFO). Once the buffer of size $B$ is full, we apply the entropy maximization algorithm to maintain samples of high entropy. Each of the samples in the main memory are stored as a pair of sample and the corresponding classifier score. The classifier score $S$ is maintained as a discretised label. Only samples with a classifier score greater than a threshold $\tau$ are accepted. Then the score is discretized into range $0 \; to \; Y$ by a linear mapping function $D$. Hence each sample is represented by its corresponding label $y$  Let the samples in the auxiliary memory be represented by $\mathcal{T} =\left\{\varphi_{j}, y_{j} \mid 1 \leq j \leq B\right\}$. Where $j$ is the index of the sample in the auxiliary memory. $y$ ranges from $[0 \; to \; Y]$.

To train a model online, buffer  $\mathcal{B}=\left\{\varphi_{k}, y_{k}\right\}_{k=1}^{B}$. This buffer is used on the rehearsal of previously seen samples. When an individual input output pair ${\varphi_k, y_k}$ arrives, a suitable algorithm decides which old sample will be replaced from the buffer. During run-time, samples from the main and auxiliary buffers are sampled to train the model. For the model to avoid catastrophic forgetting, the auxiliary memory samples must be as diverse as possible. Inspired by~\cite{entropyss}, we consider the problem of selecting a sample for rehearsal as a joint distribution of $P(\mathbf{\varphi}, Y): \mathbb{R}^{S} \times \mathbb{Z} \rightarrow[0,1]$ for random variables $\mathbf{\varphi}$ and $Y$.  The Shannon entropy of $\varphi$ and $Y$ given by, $H(\mathbf{\varphi}, Y)=-\mathbb{E}[\log P(\mathbf{\varphi}, Y)]$. Since $Y$ is a discretized classifier score for the sample, it can be considered a representation of different appearances of the sample. If $P(\varphi, y)$ becomes more predictable, then the entropy will tend to zero. We aim to make $\varphi$ as diverse as possible, which is similar to maximum entropy sampling. Shannon entropy can be re-written as:
\begin{equation}
\begin{aligned}
H(\mathbf{\varphi}, Y) &=-\mathbb{E}[\log P(\mathbf{\varphi} \mid Y)]-\mathbb{E}[\log P(Y)] \\
&=H(\mathbf{\varphi} \mid Y)+H(Y) .
\end{aligned}
\label{eqn:1}
\end{equation}

\begin{algorithm}
\caption{Entropy based sample selection with discrete classifier scores}\label{alg:algo}
\begin{algorithmic}
\Require $\varphi, S$, Auxiliary Buffer $\mathcal{A}$, $B$
\Ensure Updated $\mathcal{A}$, $BA$ training samples

\If{$\mathcal{A} < B$}
    \State $y \gets D(S)$ \Comment{Discretize classifier score to get a label}
    \State $\mathcal{A} \gets \varphi,y$
    \Comment{Add to buffer until full }
\Else
    \State $\mathcal{C} \gets$ samples of majority label in $\mathcal{A}$
    \For{$\varphi \in \mathcal{C}$} 
    
         \State $d_{i} \gets \min_{\mathbf{x}_{j}} \in \mathcal{C}\left\|\mathbf{x}_{i}-\mathbf{x}_{j}\right\|_{2}$
         
       \EndFor
    
    \State $i \sim P(i) \gets \left(1-d_{i}\right) / \sum_{j}\left(1-d_{j}\right)$
    \State $\mathbf{x}_{i}, y_{i} \leftarrow \mathbf{x}, y$
    
    \State $BA \leftarrow B + n~samples~from~\mathcal{A}$
    
\EndIf

\end{algorithmic}
\end{algorithm}
 It can be observed from Eqn~\ref{eqn:1} that the overall entropy depends on $P(Y)$ and $P(\varphi\mid Y)$. $P(Y)$ can be minimised by keeping the most common label as a criterion to replace the sample. While among the samples that fall into the same labels, there could be many repetitions of similar examples given a label. $P(\varphi\mid Y)$ is related to examples within the same labels, and it can be minimised by estimating the distribution of $P(\varphi\mid Y)$. Similar to ~\cite{entropyss} we use Kernel Density Estimation (KDE) to estimate $P(\varphi\mid Y)$, defined by:
 \begin{equation}
\begin{aligned}
 P(\mathbf{x} \mid y) \approx \frac{1}{M_{y}} \sum_{k=1}^{M_{y}} ,  K\left(\mathbf{x}-\mathbf{x}_{y}[k]\right)
\end{aligned}
\label{eqn:2}
\end{equation}
where $M_y$, is the number of examples with label $y$, and $K: \mathbb{R}^{S} \rightarrow[0,1]$ is a kernel function. Hence, given these approximations an algorithm to maximize the joint entropy $H(\mathbf{\varphi} \mid Y)$ is needed. First, we maximize $H(Y)$ by maintaining an equal number of samples with labels $Y$ across all classes. This is achieved by identifying the majority label $Y$ and then replacing the sample in the majority class with a probability $\left(1-d_{i}\right) / \sum_{j}\left(1-d_{j}\right)$ where $d_i$ is the minimum distance of $x_i$ to all the examples of the same class in the buffer. The complete process is detailed in Algorithm 1. 

We summarize our proposed method as follows. When the auxiliary buffer is not full, we keep adding all new samples that exit from the main memory until the auxiliary buffer is full. Once the buffer is full, we use the relative frequency of the class score discretised label to select the majority label. We choose an example with this label for replacement where the probability of a particular example to be replaced is high if the minimum distance to all other examples of the same label in the buffer is small.

During tracking, all the samples from main memory and a certain fixed set of $n$ samples from auxiliary memory are randomly sampled to create a $BA$ batch of samples for training the classifier. The main memory ensures current representation is learnt while the auxiliary samples ensure catastrophic forgetting is avoided.

\section{Results and Discussion}

\subsection{Datasets:}
We evaluate trackers on four well known object tracking datasets. Two of them, UAV123~\cite{uavdata} and OTB-100~\cite{otb}, are composed of short tracking videos, while LaSOT~\cite{lasot} and TrackingNet~\cite{trackingnet} are large scale datasets. The datasets are described below.
\begin{itemize}
    \item \textbf{OTB-100:} OTB dataset is a classical benchmark in Visual Object tracking containing a total of 100 videos with an average of 590 frames per video. We report the results on OTB-2015 and it is known to have tended to saturation over the years~\cite{stmtrack}. 
    \item \textbf{UAV123:} This dataset consists of 123 low altitude aerial videos captured from UAV. It is a challenging dataset due to small objects, fast motion as well as the presence of distractor objects.
    \item \textbf{LaSOT:} LaSOT is a high-quality benchmark for Large-scale Single Object Tracking. LaSOT consists of a total of 280 test videos. The videos are longer compared to other datasets with an average of 2500 frames per video in the dataset. Thus online model adaptation and without catastrophic forgetting are crucial for this dataset.
    \item \textbf{TrackingNet:} TrackingNet is a large-scale tracking dataset consisting of videos in the wild. It has a total of 30,643 videos split into 30,132 training videos and 511 testing videos, with an average of 470,9 frames.
\end{itemize}    

\subsection{Experimental Setup:}
Our proposed method is integrated in state-of-art DiMP~\cite{dimp}, PrDiMP~\cite{prdimp}, and TrDiMP~\cite{transdimp} trackers, all online learning based trackers. These trackers have been selected for evaluation because DiMP is one of the first successful real-time online learning tracker while PrDiMP and TrDiMP are improvisations on DiMP. TrDiMP is a the state-of-the-art models for real time tracking. Both PrDiMP and TrDiMP trackers are accurate on recent visual object tracking challenge datasets, and hence they have been selected for our study. We use the pre-trained models provided by the authors, and do not train the tracker further, since our proposed method is an online sample selection algorithm during that is applied during run time. We use Page-Hinckeley~\cite{agrahari2021concept} test for change detection, to decide when to update the classifier. Although we could effectively use other statistical change detection algorithms we use Page-Hinckeley as it is a sequential analysis methods for change detection (see Section 2). Additionally, it can work with continues values such as classifier scores. We set a memory buffer size of 50 similar to~\cite{dimp,prdimp} and auxiliary memory buffer as 50. The threshold $\lambda$ for change detection is set to 0.15 empirically.

In this paper, trackers are evaluated with Area Under Curve metric (AUC) of the success plot~\cite{otb}, which measures the overall performance of the tracker, under different tracked to ground truth overlap acceptance threshold. 


\subsection{Ablation Studies:}
In this subsection, we characterize our proposed method on LaSOT dataset~\cite{lasot}. In this part of our experiments, we analyze the effectiveness of different components of our proposed method and study the change detection threshold $\lambda$. 
Table~\ref{tab:abl} shows the ablation study performed for the proposed method on LaSOT~\cite{lasot} dataset. The study was performed to study the effectiveness of different components of the proposed model. The first part of the table studies the effect of change detection for model update. The samples for the main memory are collected all the frames in a circular buffer keeping a maximum of 50 samples from past. The classifier is updated only when a change in the concept is detected by statistically analysing the classifier scores for previously seen samples as a time-series data. This is performed by the change detector. This indicates that the object appearance is changing and this would be a good time to train the classifier online using the samples from the main memory. 
\begin{table}[t!]
3\caption{Ablation study: AUC accuracy of different components of our proposed method integrated in DiMP on the LaSOT dataset.}
\begin{center}
{
\begin{tabular}{|ll|}
\hline
\multicolumn{1}{|l|}{\textbf{Tracker}} & \textbf{\begin{tabular}[c]{@{}l@{}}AUC \end{tabular}} \\ \hline \hline
\multicolumn{2}{|l|}{\textbf{$\cdot$ Change Detector}} \\ \hline
\multicolumn{1}{|l|}{DiMP~\cite{dimp}} & 57.1 \\ \hline

\multicolumn{1}{|l|}{DiMP Random} & 55.4 \\ \hline
\multicolumn{1}{|l|}{DiMP Periodic 5} & 56.8 \\ \hline
\multicolumn{1}{|l|}{DiMP Periodic 10} & 56.5 \\ \hline
\multicolumn{1}{|l|}{DiMP Periodic 15} & 55.6 \\ \hline
\multicolumn{1}{|l|}{DiMP with CD} & 57.8 \\ \hline
\multicolumn{2}{|l|}{\textbf{$\cdot$ Auxiliary Memory}} \\ \hline
\multicolumn{1}{|l|}{DiMP with CD + Random Replacement} & 57.9 \\ \hline
\multicolumn{1}{|l|}{DiMP with CD + Density Replacement} & 58.2 \\ \hline
\multicolumn{1}{|l|}{DiMP with CD + Class. Score Discretised} & 58.6 \\ \hline
\end{tabular}
}
\end{center}
\label{tab:abl}
\end{table}

For the ablation study, different experiments were performed where the classifier was trained with samples in memory online randomly as indicated by "DiMP+Random" in the table. Then further experiments were performed by training the classifier online periodically as indicated by "Dimp+Periodic" every 5, 10 and 15 frames of the incoming image stream. Then finally, change detection was applied on the classifier's output scores to train the classifier online when changes were detected. It can be observed that change detection triggered classifier training outperforms all the other strategies. This is because we rely solely on the classifier score to detect concept drift and hence we train the model only on need basis. All other methods use a strategy that does not really takes classifier response into account. Hence, change detection for model update can be a efficient method for knowing when to update the tracker model given noisy tracker localisation.
\begin{table}[h]
\caption{AUC accuracy of our proposed method integrated in DiMP50 and PrDiMP50  trackers on the UAV123, OTB-100, LaSOT and TrackingNet datasets.}
\begin{center}
\scalebox{0.97}
{
\begin{tabular}{|lcccc|}
\hline

\textbf{Tracker} & \textbf{UAV123} & \textbf{OTB100} & \textbf{TrackingNet} & \textbf{LaSOT} \\ \hline \hline
\multicolumn{1}{|l|}{ECO~\cite{eco}} & 53.2 & 69.1 & - & - \\
\multicolumn{1}{|l|}{ATOM~\cite{atom}} & 63.2 & 66.4 & 70.3 & 51.5 \\
\multicolumn{1}{|l|}{MDNet~\cite{nam2016mdnet}} & - & 67.8 & - & 39.7 \\
\multicolumn{1}{|l|}{SiamRPN++~~\cite{siamrpnpp}} & - & 69.6 & 73.3 & 49.6 \\ \hline
\multicolumn{1}{|l|}{DiMP50~\cite{dimp}} & 65.3 & 68.4 & 74 & 56.9 \\
\multicolumn{1}{|l|}{ADiMP50(Ours)} & 67.6 & 69.1 & 75.5 & 58.6 \\ \hline 
\multicolumn{1}{|l|}{PrDiMP50~\cite{prdimp}} & 66.7 & 69.6 & 75.8 & 59.8 \\
\multicolumn{1}{|l|}{APrDiMP50(Ours)} & 67.8 & 70.9 & 77.1 & 61.5 \\ \hline
\multicolumn{1}{|l|}{TrDiMP~\cite{transdimp}} & 67.3 & 70.6 & - & 63.0 \\
\multicolumn{1}{|l|}{ATrDiMP(Ours)} & 68.1 & 71.5 & - & 69.0 \\ \hline
\end{tabular}

}
\end{center}
\label{tab:uav}
\end{table}

The next part of the table contains results of experiments performed to evaluate the effectiveness of our proposed classifier score discretization for maintaining with high variance in memory. As discussed in the algorithm, our proposed method replaces old samples from memory when new samples arrive. The samples to be replaced are determined by different criteria, i.e., a discrete label based on classifier scores and density of the samples within the same bin of label. First, we randomly replace samples from memory when a new sample arrives, indicated by "DiMP+CD+Random replacement". Then we use density to replace samples considering all samples under one label, "DiMP+CD+Density replacement". Finally, we segregate samples in individual bins determined by discretizing classifier scores and assigning discrete values to them. Then a density-based replacement is applied for samples within bins indicated by "DiMP+CD+Class. Score discretized" in the Table. Our proposed classifier score based discretization improves the tracker's overall AUC, indicating that the proposed method stores samples such that the samples in the memory have a high variance. This ensures to hold most information within a budgeted memory.  

\begin{figure*}
\centering
\begin{subfigure}{0.4\textwidth}
    \includegraphics[width=\textwidth]{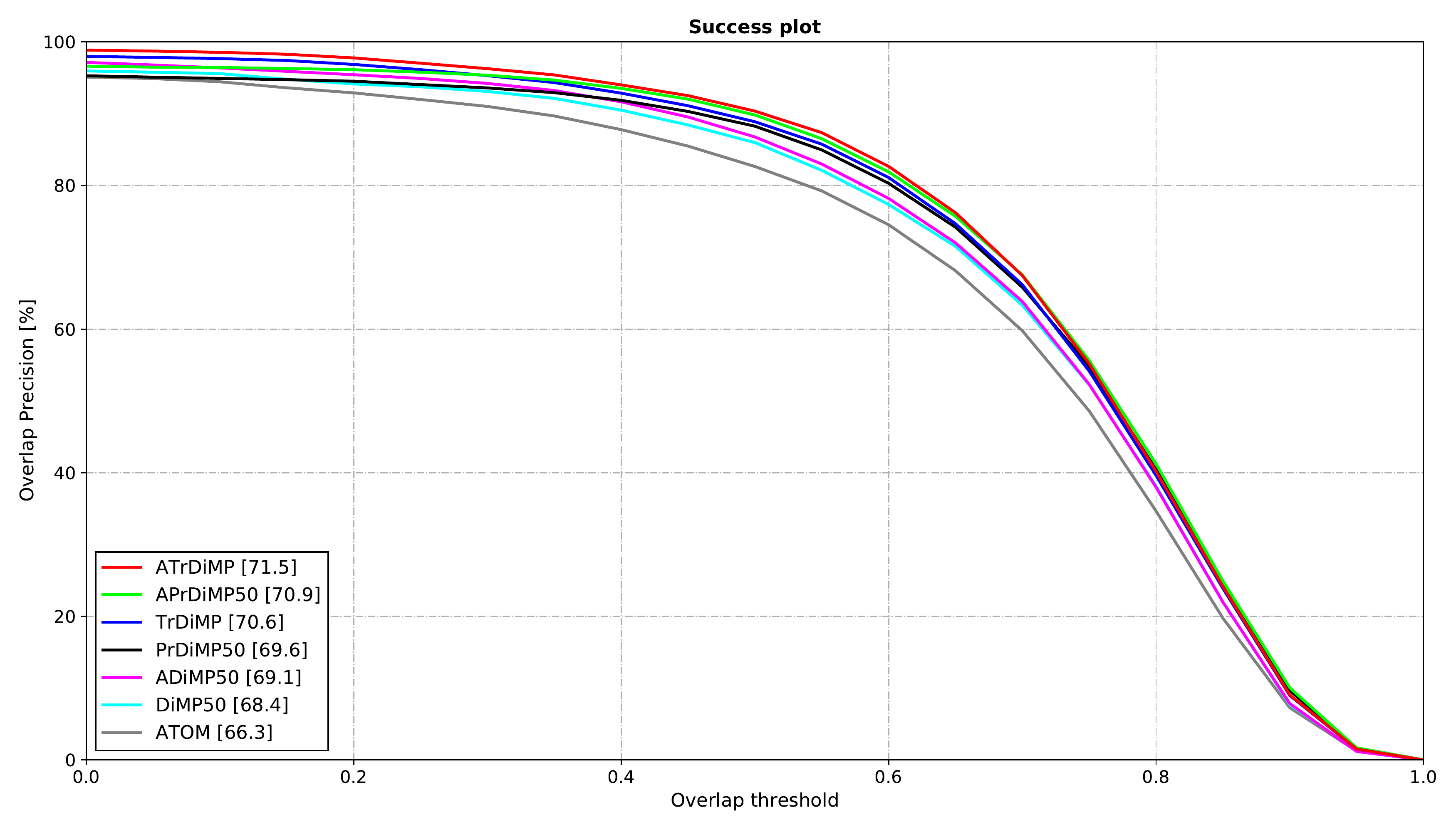}
    \caption{Success plot for OTB dataset.}
    \label{fig:first}
\end{subfigure}
\hfill
\begin{subfigure}{0.4\textwidth}
    \includegraphics[width=\textwidth]{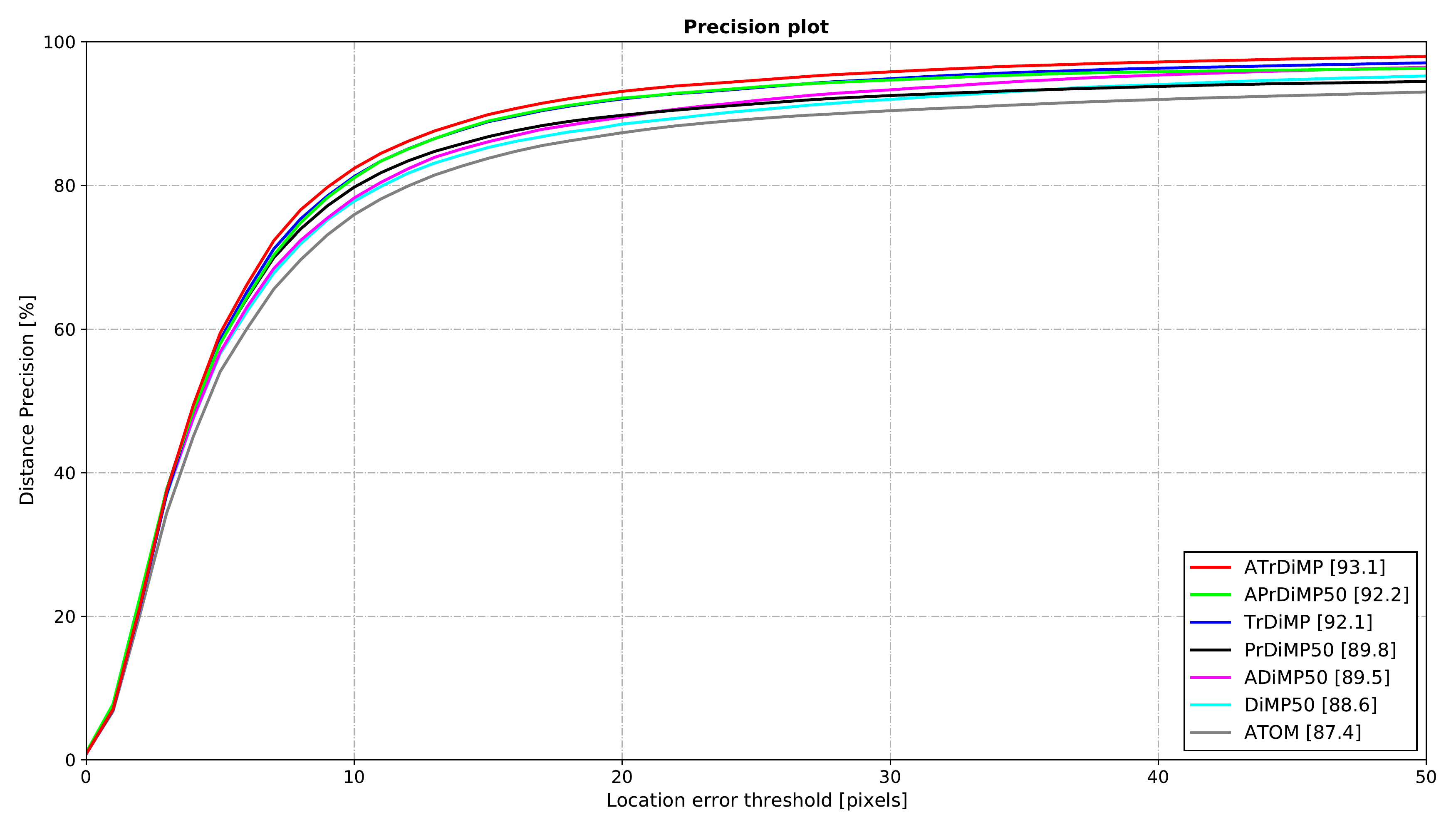}
    \caption{Precision plot for OTB dataset.}
    \label{fig:second}
\end{subfigure}
\hfill
\begin{subfigure}{0.4\textwidth}
    \includegraphics[width=\textwidth]{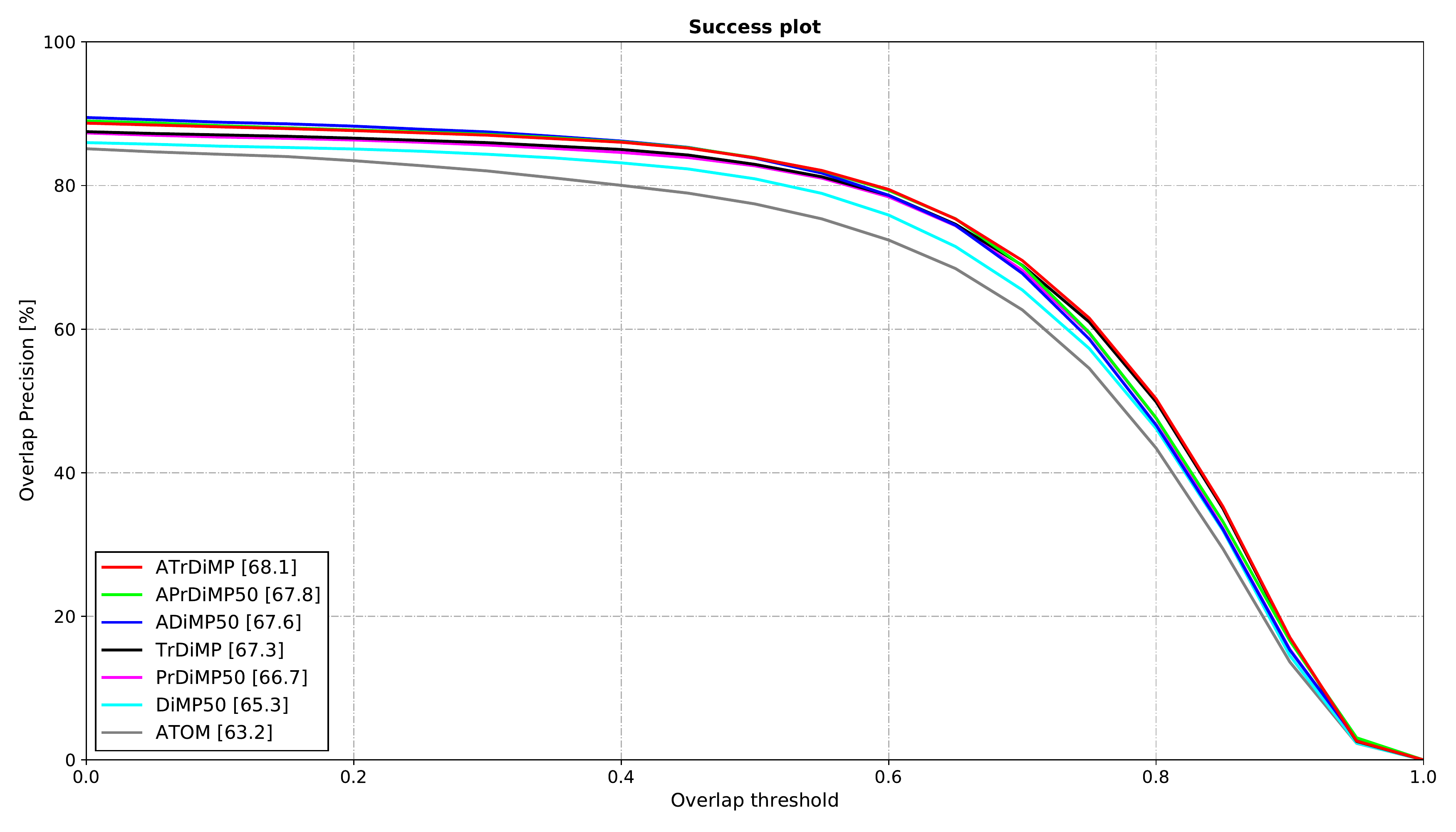}
    \caption{Success plot for UAV dataset.}
    \label{fig:third}
\end{subfigure}
\hfill
\begin{subfigure}{0.4\textwidth}
    \includegraphics[width=\textwidth]{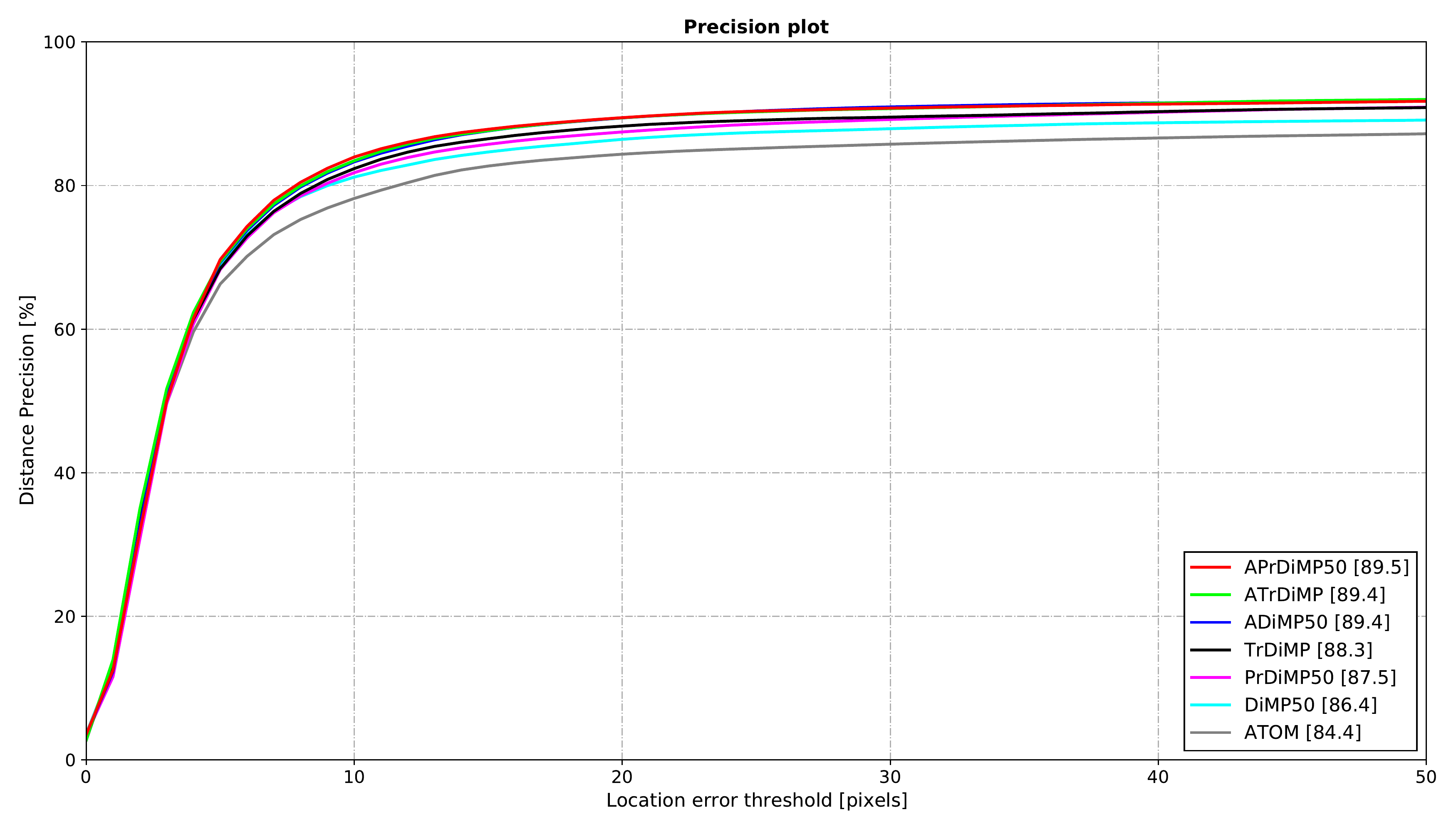}
    \caption{Precision plot for UAV dataset.}
    \label{fig:third}
\end{subfigure}
        
\caption{Success and Precision plots to compare our proposed methods with the state-of-the-art methods}
\label{fig:success}

\end{figure*}

\subsection{Integration in State-of-art Trackers:}

Our proposed approach is applied on DiMP and PrDiMP trackers, indicated by ADiMP and APrDiMP, respectively.  We compare our method with other template matching based trackers as well as other adaptive siamese tracking based trackers (Table~\ref{tab:uav}). We can observe that our method significantly improves the AUC accuracy of both state-of-the-art DiMP and PrDiMP trackers. In particular, the results with the UAV dataset appear better than that with the OTB dataset because UAV dataset videos are relatively longer than those of OTB-100. This is because longer videos have enough samples to improve the quality of samples with the auxiliary memory. 

We also evaluate of our proposed method on these datasets similar to the previous section.The Area Under Curve was chosen as the comparison metric. Our proposed ADiMP and PrDiMP trackers outperform the corresponding baselines. ADiMP achieves an improvement of 1.6\% improvement over the baseline and PrDiMP achieves an improvement of 1.7\% on the LaSOT dataset. Similarly on the TrackingNet dataset they achieve 1.5\% and  1.2\% respectively.

\subsection{Success and Precision plots:}
Fig.~\ref{fig:success} shows the comparison of success plots and precision plots of our proposed method along with state-of-the-art methods. We can observe from the plots that we achieve better AUC score in comparison with methods without our proposed change detection and sample selection. PrDiMP and TrDiMP are relative improvements over DiMP tracker with significant architectural changes or changes with additional loss functions. But Our proposed method achives a similar improvement with a simple computationally efficient technique as change detection based online learning and entropy based sample selection.

\subsection{Tracking Frame Rate}
We compare the frame rate of our tracker with other online learning methods in Table~\ref{tab:fps}. The table shows that our proposed ADiMP and APrDiMP methods run faster than the baselines by 6 and 5 FPS, respectively. This performance improvement occurs because, different than the baseline methods, we do not train during every frame of the tracked object. Our adaptive framework only updates the tracker in the presence of concept drift, significantly reducing the overall number of times the tracker was trained in a  given video. Thus, reducing the computational complexity of the system. In applications such as video surveillance, the object tracker needs to perform at real-time to avoid missing important events. Additionally longer tracking is an other important aspect of video surveillance applications specially in vehicle or person tracking to maintain unique identities during tracking. The trackers were evaluated on a Linux server with GTX1080 graphics card.

\begin{table}[]
\caption{Comparison of tracking speed of online learning trackers with our proposed method evaluated on LaSOT dataset.}
\begin{center}
{
\begin{tabular}{|l|c|}
\hline
Trackers & Frames/Second \\ \hline
ECO~\cite{eco} & 60 \\
SiamRPN++~\cite{siamrpnpp} & 35 \\
MDNet~\cite{nam2016mdnet} & 3 \\
ATOM~\cite{atom} & 30 \\
DiMP~\cite{dimp} & 38 \\
ADiMP(ours) & 44 \\ \hline
PrDiMP~\cite{prdimp} & 28 \\
APrDiMP(ours) & 33 \\ \hline
TrDiMP~\cite{transdimp} & 25 \\
ATrDiMP(ours) & 29 \\ \hline
\end{tabular}
}
\end{center}
\label{tab:fps}
\end{table}

\section{Conclusion}

We propose an online learning framework with concept drift triggered online classifier training for adaptive Siamese tracking methods. We hypothesize that in online learning of object tracking model, learning is necessary only when a concept drift is detected. Additionally we identify catastrophic forgetting in online tracking and propose entropy based sample selection for online learning in trackers.Our proposed method could be adapted into several state-of-the-art adaptive tracking methods. We evaluate our proposed method on DiMP, PrDiMP and TrDimp trackers on OTB, UAV, LaSOT and TrackingNet datasets. Our approach improves the performance of state-of-the-art adaptive tracking methods by an average of 2\% in AUC measure in addition to improving the overall tracking speed.

\FloatBarrier
\bibliographystyle{sort}
\bibliography{egbib}


\end{document}